\pdfoutput=1

\documentclass[preprint,12pt]{elsarticle}

\usepackage{amssymb}
\usepackage{amsmath}
\usepackage{xcolor}
\usepackage[table,xcdraw]{xcolor}
\usepackage{algorithm}
\usepackage{algorithmic}
\usepackage{multirow}
\usepackage{url}
\journal{Nuclear Physics B}

\begin{document}

\begin{frontmatter}



\title{A fairness-aware extension of Stochastic Multicriteria Acceptability Analysis for ranking}


\author{Guilherme Dean Pelegrina, Renata Pelissari} %% Author name

\affiliation{organization={Engineering School, Mackenzie Presbyterian University},%Department and Organization
            addressline={930 Consolação St}, 
            city={São Paulo},
            postcode={01302-907}, 
            state={São Paulo},
            country={Brazil}}

%\author{Renata Pelissari} %% Author name

%\affiliation{organization={Engineering School, Mackenzie Presbyterian University},%Department and Organization
%            addressline={930 Consolação St}, 
%            city={São Paulo},
%            postcode={01302-907}, 
%            state={São Paulo},
%            country={Brazil}}

%% Abstract
\begin{abstract}
 Fairness has become a central concern in ranking problems involving individuals or social groups, particularly under the Responsible Artificial Intelligence agenda. In Multi-Criteria Decision Analysis, Stochastic Multicriteria Acceptability Analysis (SMAA) provides a robust framework for handling uncertainty and incomplete preference information, but it does not explicitly address fairness in the resulting rankings. This paper proposes SMAA-Fair, a fairness-aware extension of SMAA for ranking problems. The approach reweights the simulated rankings generated by SMAA according to their level of group fairness, so that fairer rankings contribute more strongly to the acceptability indices and central weights vector. The framework is independent of the aggregation model and can incorporate different fairness metrics. In this study, Statistical Parity, normalized discounted Kullback--Leibler divergence (rKL) and normalized discounted cumulative Kullback--Leibler divergence (nDKL) are adopted. Rankings are derived from the fairness-adjusted acceptability matrix using expected ranking and maximum acceptability ranking. We also derive the central weight according to the degree of fairness in the obtained rankings. Numerical experiments with synthetic and real data show that SMAA-Fair improves the representation of protected groups among favourable ranking positions, while preserving robustness to preference uncertainty.
\end{abstract}



%% Keywords

\end{frontmatter}

\section{Introduction}
    
    The increasing use of Artificial Intelligence (AI) raises sustainability challenges that go beyond environmental concerns, including social aspects such as equity, transparency, accountability, and fairness. Among these, fairness is a key element of the Responsible AI agenda, aiming to ensure that algorithms operate equitably and inclusively \citep{Chen2023SurveyAISustainability}.

    Several notions of fairness have been discussed in the literature, including distributive, procedural, restorative, and retributive fairness \citep{Varshney2022TrustworthyMachineLearning}.
    Distributive fairness concerns equality in the outcomes people receive, while procedural justice concerns equality in the processes through which such outcomes are determined. In contrast, restorative fairness addresses the repair of harm caused by past decisions, while
    retributive fairness relates to sanctioning wrongdoing.

    In the context of algorithmic fairness, fairness considerations are typically framed in terms of distributive fairness, since algorithms are primarily designed to allocate resources or opportunities and to produce outcomes \citep{Mitchell2021}. Consequently, fairness analysis often examines whether different individuals or groups receive comparable results in applications such as credit approval \citep{Kozodoi2022,ZHANG2024121484,Carrizosa2026}, healthcare eligibility \citep{grote2020ethics, chen2023algorithmicfairness}, or candidate selection \citep{ZHANG2022101848}. In this context, the notion of groups plays a central role \citep{Mitchell2021, Varshney2022TrustworthyMachineLearning}. At the individual level, groups are usually defined according to protected attributes such as race, gender, age, or disability status, as established by legal or regulatory frameworks in each application domain. Based on these attributes, individuals are commonly categorized into privileged and unprivileged groups, reflecting historical patterns of advantage or disadvantage. However, fairness analysis is not limited by groups of individuals. One may also evaluate fairness, for instance, in terms of groups of countries~\citep{rocha2025fair} or universities~\citep{Kostreva2004,Radovanovi2022}. Fairness assessment, therefore, relies on comparing outcomes across these groups to detect and mitigate disparities \citep{Mitchell2021}. 

    %Unfairness algorithmic can emerge from two main sources: biases in data and algorithmic biases \citep{Mehrabi2022BiasFairnessSurvey}. Bias in data often occurs when datasets contain imbalanced or unrepresentative information, while algorithmic bias stems from the assumptions and design choices embedded in the models. Promoting fair algorithms involves identifying and eliminating such biases, ensuring that AI systems treat different individuals and groups equitably. 

   To quantify such disparities, a wide range of fairness metrics has been proposed, each tailored to specific decision problems, such as classification or ranking \citep{Garg2020, deho2022existing}. In ranking problems, which are the focus of this paper, fairness metrics are designed to ensure equitable representation within the most influential positions of the ranking. A common operationalization requires that the proportions of individuals from privileged and unprivileged groups appearing in the top-$k$ positions reflect their proportions in the underlying population or dataset \citep{Yang2017, zehlike2017, Celis2018, singh2018fairness}.

   Although fairness in ranking has been extensively studied in machine learning, particularly in applications such as recommendation and search systems
   \citep{Wang2023, geyik2019fairness}, similar concerns also arise in Multi-Criteria Decision Analysis (MCDA), where alternatives are evaluated with respect to multiple, potentially conflicting criteria \citep{greco2019}. For instance, in the case study reported in \citep{govindan2017employee}, criteria such as productivity, technical competence, interpersonal skills, leadership potential, and attendance are aggregated to rank employees for promotions, bonuses, or training opportunities. While such criteria may appear objective, some components rely on subjective judgments or reflect differences in working conditions and career opportunities. As a result, MCDA applications involving individuals may inadvertently reproduce or amplify existing disparities.

   Fairness in MCDA remains a recent research topic, with most scientific studies having been conducted over the last decade~\citep{Castillo2019, Pitoura2022, Yang2017, Ghazimatin2022, Celis2018, singh2018fairness, rocha2025fair}. However, incorporating fairness considerations into MCDA when preferences or model parameters are uncertain or only partially specified remains an open research area. Since such uncertainty is common in real decision-making contexts \citep{greco2019, pelissari2019}, there is a natural motivation to combine fairness-aware MCDA with methods that account for uncertainty.
	
    %Fairness in MCDA remains a relatively recent research topic, with only a limited number of studies explicitly addressing fairness metrics and fairness-aware modeling approaches. For instance, the study \citep{rocha2025fair} developed a fairness-oriented MCDA framework in which the parameters of additive and non-additive aggregation models are optimized to reduce disparities in group outcomes, demonstrating how fairness metrics can be directly embedded into the decision-making process. 
    
    %While incorporating fairness metrics into MCDA models represents an important advance, the robustness of fairness-related conclusions remains an open issue when preference information or model parameters are uncertain or partially specified. Since such uncertainty is common in real decision-making contexts \citep{greco2019, pelissari2019}, there is a natural motivation to combine fairness-aware MCDA with methods that account for uncertainty.

    Stochastic Multicriteria Acceptability Analysis (SMAA) provides a well-established framework for addressing uncertainty and incomplete preference information in MCDA \citep{lahdelma2001smaa2}. In SMAA, the relative importance of criteria is not assumed to be precisely known: criteria weights may be partially specified, imprecise, or even completely missing. Instead of relying on a single deterministic preference model, SMAA explores the space of admissible preference parameters through Monte Carlo simulation. At each iteration, a feasible set of parameters is randomly sampled, and a chosen aggregation model is applied to produce a complete ranking of the alternatives. By repeating this process over a large number of simulations, SMAA yields a probabilistic description of the decision problem. In ranking problems, the main descriptive measure is the rank acceptability index, which represents the probability that an alternative to be ranked in each position of the ranking. Another descpritive measure given by SMAA in ranking problems is the central weight vector that represents the average weight vector among those leading a given alternative to the first position in the ranking.

    This paper proposes a fairness-aware extension of SMAA, referred to as {SMAA-Fair}, which integrates group fairness considerations into the stochastic evaluation process. The proposed framework modifies how simulated rankings contribute to the rank acceptability index, giving more importance to rankings that exhibit higher levels of fairness. The central weight vector is also modified such that the center of mass shifts toward regions associated with fairer rankings.
    
    Although the framework is flexible with respect to the choice of fairness metric applied, in this study, we adopted measures based on the notion of Statistical Parity, normalized discounted Kullback--Leibler divergence (rKL) \citep{Yang2017} and normalized discounted cumulative Kullback--Leibler divergence (nDKL) \citep{geyik2019fairness}. Since the outcomes of SMAA-Fair remain probabilistic, a single representative ranking is derived using both expected ranking~\citep{lahdelma2001smaa2} and maximum acceptability ranking~\citep{vetschera2017deriving} methods to support numerical evaluation and comparative analysis. Experiments on both synthetic and real datasets are then used to show that the proposed framework effectively improves fairness under preference uncertainty.
    
    The remainder of the paper is organized as follows. Section~\ref{sec:back} presents the theoretical background necessary to support the development of our proposal. Section \ref{sec:prop} introduces the proposed SMAA-Fair method for ranking. Section 4 reports numerical experiments in both synthetic and real datasets. Finally, Section 5 concludes the paper. 

\section{Background}
\label{sec:back}

    This section presents the theoretical foundations underlying the proposed SMAA-Fair
    framework. We first introduce the basic principles of Stochastic Multicriteria
    Acceptability Analysis (SMAA) in the context of ranking problems. We then review
    key concepts and metrics related to fairness in ranking.

\subsection{MCDA and SMAA for ranking}
\label{subsec:mcda_smaa}

    We consider a MCDA ranking problem in which a finite set of alternatives
    \( \mathcal{A} = \{a_1, a_2, \dots, a_m\} \) is evaluated with respect to a set of criteria
    \( C = \{c_1, c_2, \dots, c_n\} \). Each criterion \( c_j \) is associated with an
    evaluation function \( q_j: \mathcal{A} \to \mathbb{R} \), assigning a performance score to
    each alternative. Usually, the criteria evaluations are represented by a matrix
    \begin{equation}
        M =
        \begin{bmatrix}
            q_1(a_1) & q_2(a_1) & \cdots & q_n(a_1) \\
            q_1(a_2) & q_2(a_2) & \cdots & q_n(a_2) \\
            \vdots & \vdots & \ddots & \vdots \\
            q_1(a_m) & q_2(a_m) & \cdots & q_n(a_m)
        \end{bmatrix}
    \end{equation}
   in which $q_j(a_i)$ indicates the evaluation of alternative $a_i$ on criterion $c_j$. In classical MCDA, preferences over alternatives are obtained by
    aggregating these criteria evaluations through a preference model. A common approach consists in defining an aggregation function
    \[
    u: \mathbb{R}^n \times W \to \mathbb{R},
    \]
    \noindent where \( W \) denotes the space of feasible weight vectors
    \( \mathbf{w} = (w_1, \dots, w_n) \), which represent the relative importance of the criteria.
    For a given weight vector \( \mathbf{w} \in W \), the aggregated score of alternative
    \( a_i \), $i=1, \ldots, m$, is given by \( u(a_i, \mathbf{w}) \). The resulting ranking is defined as a mapping
    \[
    \pi: \mathcal{A} \to \{1, 2, \dots, m\},
    \]
    \noindent such that alternatives with higher aggregated scores are assigned better positions,
    that is,
    \[
    u(a_i, \mathbf{w}) \geq u(a_{i'}, \mathbf{w}) \;\Rightarrow\; \pi(a_i) \leq \pi(a_{i'}).
    \]
    
    However, in practice, decision problems rarely rely on complete and precise preference information. Specially criteria weights may be only partially specified, uncertain, or even completely unknown. The Stochastic Multicriteria Acceptability Analysis (SMAA) methodology was developed to address such situations \citep{lahdelma1998smaa,lahdelma2001smaa2}.

    Within the SMAA framework, uncertainty in the decision maker's preferences is propagated through repeated simulations. At each iteration $l = 1,\dots,L$, a criteria weight vector is randomly sampled from the feasible weight space $W$. Given the sampled parameters, the alternatives are evaluated using a multicriteria aggregation model, yielding a complete and deterministic ranking of the alternatives.

    We denote by
    \[
    \pi^{(l)} : \mathcal{A} \rightarrow \{1,\dots,m\}
    \]
    \noindent the ranking obtained at iteration $l$, where $\pi^{(l)}(a_i)$ represents the position assigned to alternative $a_i$ under the sampled preference parameters.

    By aggregating the results of a large number of iterations, SMAA produces
    probabilistic description measures of the decision problem. Depending on the specific SMAA variant, different measures can be derived to describe the relative performance and robustness of the alternatives across the simulated scenarios. In the classical SMAA-2 formulation, three measures are central: the rank acceptability index, the central weight vector, and the confidence factor. The rank acceptability index, denoted by \( b_i^s \), represents the probability that alternative \( a_i \) attains the $s$-th rank position. This value lies in \([0,1]\), with values closer to 1 indicating strong robustness of the alternative at a specific rank. This index may be represented by the matrix $B$ defined by

    \begin{equation}\label{eq_rank_accept_matrix}
        B =
        \begin{bmatrix}
            b_1^1 & b_1^2 & \cdots & b_1^m \\
            b_2^1 & b_2^2 & \cdots & b_2^m \\
            \vdots & \vdots & \ddots & \vdots \\
            b_m^1 & b_m^2 & \cdots & b_m^m
        \end{bmatrix}.
    \end{equation}

    The central weight vector $\mathbf{w}_i^c$ characterizes the preference structure that makes alternative $a_i$ ranks first. Formally, for an alternative $a_i$, the central weight vector is defined as the expected value of the weight vectors conditioned on the event that $a_i$ attains the first position. Through the Monte Carlo simulation, $\mathbf{w}_i^c$ can be estimated as follows:

    \begin{equation}
        \mathbf{w}_i^c = \frac{1}{|W_i|}\sum_{\mathbf{w} \in W_i}\mathbf{w},
    \end{equation}
    where $W_i$ denotes the subset of sampled weights, with cardinality $|W_i|$, leading to the first-rank assignment of alternative $a_i$. Therefore, $\mathbf{w}_i^c$ represents the barycenter of the subset of preference parameters that support the selection of alternative $a_i$. Presenting these vectors to decision-makers enables an inverse perspective: rather than starting from weights to compute rankings, the analysis reveals which weights would justify each alternative being ranked first. 
    
    The confidence factor $p_i^c$ complements this interpretation, measuring the probability that alternative $a_i$ actually ranks first when evaluated with its central weight vector.  

    While SMAA primarily provides probabilistic characterizations, in many applications it is necessary to summarize the results into a single ranking. Two common approaches to achieve that are based on the expected ranking \citep{lahdelma2001smaa2} and the maximum acceptability ranking~\citep{vetschera2017deriving}.
    
    The expected ranking provides a probabilistic measure of the average position an alternative occupies across multiple possible rankings. Formally, if \( b_i^s \) denotes the probability that alternative \( a_i \) is placed in position \( s \in \{1,\ldots,m\} \), its expected ranking is defined as
    \begin{equation}
    \text{ER}(a_i) = \sum_{s=1}^{m} s \cdot b_i^s.
    \end{equation}
    This quantity corresponds to the mean of the rank distribution associated with \( a_i \), giving greater weight to positions that occur more frequently. Lower expected ranking values indicate that an alternative tends to appear near the top of the rankings across the entire distribution of scenarios. Therefore, the alternatives are ranked according to the expected values in ascending order. The theoretical foundation for this interpretation originates in classical ranking theory. In \citep{mak1985interpretation}, it is shown that the rank of an item can be treated as a random variable, and that its expectation naturally summarizes the central tendency of its position across all possible permutations, weighted by their respective probabilities. This perspective aligns directly with the SMAA framework: because each SMAA iteration generates a ranking based on a sampled parameter configuration, the position of alternative \( a_i \) across iterations behaves exactly as such a random variable. Consequently, the SMAA expected ranking corresponds to the expectation of this stochastic rank variable, obtained by combining the probabilities \( b_i^s \) of the alternative occupying each position \( s \in \{1,\ldots,m\} \).

    Alternatively, the maximum acceptability ranking (MAR)~\citep{vetschera2017deriving} is a method that aims at finding a ranking that maximizes the average probability of the assigned ranks. Assume $\beta_i^s$ as a binary variable indicating that alternative $a_i$ is assigned to rank $s$. By imposing that each alternative is assigned to a single position and that each position has a single alternative, the MAR-derived ranking is obtained by solving the following optimization problem:

    \begin{align}
\max_{\beta_i^s} \quad & \sum_{i=1}^{m} \sum_{s=1}^{m} b_i^s \, \beta_i^s
\label{eq:mar_obj} \\
\text{s.t.} \quad
& \sum_{i=1}^{m} \beta_i^s = 1, 
&& \forall s = 1,\ldots,m,
\label{eq:mar_c1} \\
& \sum_{s=1}^{m} \beta_i^s = 1, 
&& \forall i = 1,\ldots,m,
\label{eq:mar_c2} \\
& \beta_i^s \in \{0,1\},
&& \forall i = 1,\ldots,m,\; s = 1,\ldots,m.
\label{eq:mar_c3}
\end{align}

    Note that MAR method does not assign a score for each alternative. The solution of~\eqref{eq:mar_c3} leads to $\beta_i^s$, for all $i,s =1, \ldots, m$, that compose the single ranking of alternatives.
    
\subsection{Fairness in ranking: Concepts and Metrics}
\label{subsec:metrics}

    The notion of fairness has been extensively examined in machine learning, where most definitions arise in the context of supervised prediction. In these settings, fairness metrics are defined with respect to the true label \(Y\). Measures such as equalized odds, equal opportunity, predictive parity and conditional parity \citep{Verma2018,Mehrabi2022BiasFairnessSurvey,zhou2022bias,bdcc7010015} assess fairness by comparing predictions \(\hat{Y}\) with true outcomes across groups defined by a protected attribute.

    Ranking problems differ fundamentally from supervised prediction tasks. Rather than producing a single outcome, a ranking yields an ordered list of individuals or items. Since higher-ranked positions are associated with greater visibility and access to opportunities, fairness in ranking is not limited to allocation itself, but also to how exposure is distributed along the ranking. This structural characteristic motivates the use of fairness notions and metrics that explicitly account for position bias and unequal representation across ranking positions.

    In the context of ranking, one widely adopted notion is group fairness. It requires that privileged and unprivileged groups appear along the ranking in proportions that reflect their prevalence in the underlying population. Applied to ordered outcomes, this principle implies that favourable ranking positions should correspond to equitable opportunities across groups. This is the perspective of fairness that we adopt in our paper. 

    A second perspective is exposure fairness, which requires that groups receive a fair allocation of exposure (\textit{i.e.}, the expected visibility determined by the positions that items occupy in the ranking and by the position-bias weights capturing user attention). Even when the proportions of each group in the ranked list are balanced, their actual visibility may still be unequal \citep{singh2018fairness}.

    A third perspective is {pairwise fairness}, which focuses on the relative ordering between individuals from different groups \citep{beutel2019fairness}. A ranking satisfies pairwise fairness when unprivileged individuals are not systematically placed below less-qualified privileged ones. This notion is particularly relevant in evaluative contexts where fairness concerns center on comparative treatment rather than aggregate representation.

    %A fourth perspective relies on {information-theoretic fairness metrics}, which quantify disparities between the distributions of outcomes assigned to each group. These metrics capture global divergence in treatment and complement the structural notions described above. For instance, the normalized Kullback--Leibler divergence (nDKL) provides a measure in \([0,1]\) of how different the group-level exposure or allocation distributions are \citep{Wang2023}. Such distributional metrics are particularly relevant when rankings induce probability distributions—such as in probabilistic ranking models or in the stochastic acceptability indices produced by MCDA methods.

    The next question is how these objectives can be operationalized within ranking systems. {Post-processing approaches} adjust a ranking after it is generated, for instance by reordering or reweighting items to meet fairness constraints. These methods allow fairness to be introduced flexibly on top of existing ranking algorithms. {In-processing approaches}, by contrast, embed fairness constraints directly within the ranking algorithm or simulation process, ensuring that the generated rankings satisfy fairness requirements but potentially restricting the feasible solution space \citep{caton2024fairness}.

    These implementation strategies are orthogonal to the three fairness perspectives discussed above. While group, exposure and pairwise fairness describe \emph{which} aspects of fairness are being evaluated, post-processing and in-processing approaches specify \emph{how and when} fairness interventions are incorporated into the ranking pipeline (either by adjusting the output or by modifying the generative mechanism that produces the ranking). 

    Having outlined the main perspectives through which fairness in ranking can be evaluated, the next step is to formalize the specific fairness metrics that will be incorporated into our proposal. As we are considering here group fairness, we focus on three measures: Statistical Parity, normalized discounted Kullback--Leibler divergence (rKL) \citep{Yang2017} and normalized discounted cumulative Kullback--Leibler divergence (nDKL) \citep{geyik2019fairness}.

    Before properly present the adopted measures, it is worth defining some elements associated with groups and ranking positions. In ranking-based decision problems, higher positions are typically interpreted as yielding more favourable outcomes. To formalize this notion, we define a subset of ranking positions $\mathcal{F}_k \subseteq \{1,\dots,m\}$, $K \leq m$, referred to as the set of $k$ favourable positions. For example, $\mathcal{F}_1 = \{1\}$ indicates the top-ranked position while $\mathcal{F}_3 = \{1,2,3\}$ represents the set of the first three positions in the ranking.

    Let $\pi : \mathcal{A} \to \{1,\dots,m\}$ denote a ranking, and let $\mathcal{G} = \{g_1, g_2\}$ denote the privileged and unprivileged groups. For each group $g_t \in \mathcal{G}$, we define
    \[
    \mathcal{G}_t = \{a_i \in A : g(a_i) = g_t\}
    \]
    \noindent as the subset of alternatives belonging to that group. Without loss of generality, let us assume that $g_1$ and $g_2$ are the privileged and unprivileged groups, respectively.

\subsubsection{Statistical Parity}

The first fairness metric adopted in this paper is the Statistical Parity. This fairness notion is widely used in classification tasks, requiring the probability of classification to be independent of sensitive attributes~\citep{Dwork2012}. In the context of ranked outputs, this concept is naturally extended to top-$k$, which compares the group proportions observed in the top-$k$ positions with their proportions in the overall population.

An alternative $a_i$ is said to receive a favourable outcome in ranking $\pi$ whenever $\pi(a_i) \in \mathcal{F}_k$. Moreover, assume

\begin{equation}
\label{eq:prob_sp}
P_{\mathrm{SP}}(\mathcal{F}_k \mid g_t)
=
\frac{1}{|\mathcal{G}_t|}
\sum_{a_i \in \mathcal{G}_t}
\mathbf{1}_{\mathcal{F}_k}\bigl(\pi(a_i) \bigr), 
\end{equation}
where $|\mathcal{G}_t|$ represents the cardinality of $\mathcal{G}_t$ and $\mathbf{1}_{\mathcal{F}_k}(\cdot)$ is an indicator function such that $\mathbf{1}_{\mathcal{F}_k}\bigl(\pi(a_i)\bigr)=1$ if $\pi(a_i) \in \mathcal{F}_k$, and 0 otherwise. The Statistical Parity (SP) for ranking $\pi$ is then defined as

\begin{equation}
\label{eq:sp}
    \mathrm{SP}
=
P_{\mathrm{SP}}(\mathcal{F}_k \mid g_2)
-
P_{\mathrm{SP}}(\mathcal{F}_k \mid g_1). 
\end{equation}

Values of $\mathrm{SP}$ close to zero indicate parity in the allocation of favourable outcomes, whereas negative values indicate that the unprivileged group is under-represented among the favourable positions.

\subsubsection{Normalized discounted Kullback--Leibler divergence}

While SP focuses on the representation of groups within a favourable position, a complementary perspective on fairness is obtained by examining how alternatives from different groups are distributed across different top-$k$ rankings. One approach based on such idea is the normalized discounted Kullback--Leibler divergence (rKL) \citep{Yang2017}, which evaluates fairness by measuring deviations from statistical parity across multiple top-$k$ rankings. The central principle is that the proportion of privileged and unprivileged groups within  multiple top-$k$ rankings should reflect their proportions in the overall population.

Consider $P_{\mathrm{KL}}(\mathcal{A} \mid g_t)$ as the proportion of alternatives from group $g_t$ among all alternatives. It can be defined as follows:

\begin{equation}
\label{eq:prob_rkl_all}
P_{\mathrm{KL}}(\mathcal{A} \mid g_t)
=
\frac{1}{|\mathcal{A}|}
\sum_{a_i \in \mathcal{G}_t}
\mathbf{1}_{\mathcal{A}}\bigl(\pi(a_i) \bigr), 
\end{equation}
where the indicator function is such that $\mathbf{1}_{\mathcal{A}}\bigl(\pi(a_i)\bigr)=1$ if $\pi(a_i) \in \mathcal{A}$, and 0 otherwise. Moreover, consider the proportion of alternatives from group $g_t$ in a top-$k$ ranking as follows:

\begin{equation}
\label{eq:prob_rkl}
P_{\mathrm{KL}}(\mathcal{F}_k \mid g_t)
=
\frac{1}{|\mathcal{F}_k|}
\sum_{a_i \in \mathcal{G}_t}
\mathbf{1}_{\mathcal{F}_k}\bigl(\pi(a_i) \bigr).
\end{equation}

For a prefix $k$, the Kullback–Leibler divergence between $P_{\mathrm{KL}}(\mathcal{F}_k \mid g_t)$ and $P_{\mathrm{KL}}(\mathcal{A} \mid g_t)$, for all groups $t$, is defined by:

\begin{equation}
\label{eq:kl_div}
    D_{\mathrm{KL}}(P_{\mathrm{KL}}(\mathcal{F}_k) \parallel P_{\mathrm{KL}}(\mathcal{A}))
=
\sum_{g_t \in \mathcal{G}}
P_{\mathrm{KL}}(\mathcal{F}_k \mid g_t) \log \frac{P_{\mathrm{KL}}(\mathcal{F}_k \mid g_t)}{P_{\mathrm{KL}}(\mathcal{A} \mid g_t)}.
\end{equation}
The rKL metric aggregates these divergences across a predefined set of prefixes $k \in \mathcal{K}$ using a logarithmic discount factor that emphasizes higher-ranked positions. It is defined by:

\begin{equation}
    \mathrm{rKL} = \frac{1}{Z} \sum_{k \in \mathcal{K}} \frac{D_{\mathrm{KL}}(P_{\mathrm{KL}}(\mathcal{F}_k) \parallel P_{\mathrm{KL}}(\mathcal{A}))}{\log_2 k},
\end{equation}
where the normalization constant $Z$ is equal to the maximum attainable value of such a measure, ensuring that $\mathrm{rKL} \in [0,1]$. For instance, the worst case, which maximize an unfair ranking, is achieved when all alternatives from the protected group appear in the last positions (or in the first positions). Assume $P_{\mathrm{KL}}^{\min} = \min_{g_t} P_{\mathrm{KL}}(\mathcal{A} \mid g_t)$. Normalizer $Z$ can be calculated as $\sum_{k \in \mathcal{K}} \frac{\log \left(1/P_{\mathrm{KL}}^{\min} \right)}{\log_2 k}$.

Therefore, in the ideal fair scenario, $\mathrm{rKL} = 0$ as the ranking satisfies statistical parity at all evaluated $k \in \mathcal{K}$. When this metric increases, the deviations from population-level group proportions become more pronounced, particularly near the top of the ranking.

\subsubsection{Normalized discounted cumulative Kullback--Leibler divergence}

The normalized discounted cumulative Kullback--Leibler divergence (nDKL) \citep{geyik2019fairness} adopts an idea related to the rKL but a distinct perspective by evaluating representation imbalance at all prefixes of the ranking. Rather than restricting attention to a predefined set $\mathcal{K}$, nDKL measures how the empirical group distribution evolves continuously along the ranking.

Assume the KL divergence $D_{\mathrm{KL}}(P_{\mathrm{KL}}(\mathcal{F}_i) \parallel P_{\mathrm{KL}}(\mathcal{A}))$ as defined in Equation~\eqref{eq:kl_div}. These prefix-level divergences are aggregated using a logarithmic discount factor, which leads to the following nDKL metric:

\begin{equation}
    \mathrm{nDKL} = \frac{1}{Z} \sum_{i=1}^{m} \frac{D_{\mathrm{KL}}(P_{\mathrm{KL}}(\mathcal{F}_i) \parallel P_{\mathrm{KL}}(\mathcal{A}))}{\log_2 \left(i+1 \right)},
\end{equation}
where the normalizer $Z = \sum_{i=1}^{m} \frac{1}{\log_2 \left(i+1 \right)}$. Therefore, nDKL is equal to zero if and only if the group distribution at every prefix matches the target distribution.

Although in both rKL and nDKL metrics assumes the same Kullback--Leibler divergence, they differ fundamentally in their modeling assumptions. The rKL metric focuses on statistical parity at selected cut-off points and is explicitly normalized by its worst-case scenario, yielding an interpretable bounded measure. In contrast, nDKL evaluates representation imbalance across all prefixes, offering a finer-grained assessment of how disparities accumulate along the ranking. The two measures are therefore complementary and jointly provide a comprehensive view of fairness in ranked outcomes.

\section{Proposed SMAA-Fair framework}
\label{sec:prop}

The proposed SMAA-Fair framework extends the classical Stochastic Multi-criteria Acceptability Analysis (SMAA) methodology for ranking by incorporating fairness-aware evaluations into the construction of the rank acceptability index and central weight vector. The central idea is to preserve the standard SMAA simulation process
while modulating the contribution of each simulated ranking according to its assessed level of group fairness. Algorithm~\ref{alg:smaa_fair} presents a pseudocode of our proposal.

\begin{algorithm}[H]
\caption{SMAA-Fair}

\begin{algorithmic}[1]
\label{alg:smaa_fair}

\REQUIRE Decision matrix $M$, group labels $\mathcal{G}$, number of iterations $L$

\STATE Normalize the decision matrix $M$
\STATE Generate $L$ feasible weight vectors $\{\mathbf{w}^{(l)}\}_{l=1}^{L}$

\FOR{$l = 1,\ldots,L$}
    \STATE Compute global scores: $u\left(\mathbf{w}^{(l)} \right)= M \mathbf{w}^{(l)}$
    
    \STATE Obtain ranking $\pi^{(l)}$
    
    \STATE Evaluate fairness metrics: $\text{SP}^{(l)}$, $\text{rKL}^{(l)}$ and $\text{nDKL}^{(l)}$
\ENDFOR

\STATE Compute worst-case fairness values: $\text{SP}_{\max} = \max_{l = 1,\ldots,L} \left( \bigl|\mathrm{SP}^{(l)}\bigr|\right)$, 
$\text{rKL}_{\max} = \max_{l = 1,\ldots,L} \left( \mathrm{rKL}^{(l)}\right)$ and $\text{nDKL}_{\max} = \max_{l = 1,\ldots,L} \left( \mathrm{nDKL}^{(l)}\right)$

\FOR{$l = 1,\ldots,L$}

    \STATE Obtain ranking $\pi^{(l)}$
    
    \FOR{each alternative $a_i$ in position $s$ of $\pi^{(l)}$}

        \STATE Update classical SMAA acceptability index: $\hat{b}_{i}^{s} \leftarrow \hat{b}_{i}^{s} + 1$
        
        \STATE Update fairness-weighted acceptability indices: $\hat{b}_{i}^{s,SP}
        \leftarrow \hat{b}_{i}^{s,SP} + (\text{SP}_{\max}- |\mathrm{SP}^{(l)}|)$, $\hat{b}_{i}^{s,rKL} \leftarrow \hat{b}_{i}^{s,rKL} + (\text{rKL}_{\max}-\mathrm{rKL}^{(l)})$ and $\hat{b}_{i}^{s,nDKL} \leftarrow \hat{b}_{i}^{s,nDKL} + (\text{nDKL}_{\max}-\mathrm{nDKL}^{(l)})$
        
    \ENDFOR

    \STATE Let $a_{i^\ast}$ denote the alternative ranked first in $\pi^{(l)}$

    \STATE Store $\mathbf{w}^{(l)}$ in the set
    $W_{i^\ast}$

    \STATE Store fairness-adjusted weights: $\omega_{SP}^{(l)} = \text{SP}_{\max}-|\mathrm{SP}^{(l)}|$,  $\omega_{rKL}^{(l)} = \text{rKL}_{\max}-\mathrm{rKL}^{(l)}$ and  $\omega_{nDKL}^{(l)} = \text{nDKL}_{\max}-\mathrm{nDKL}^{(l)}$
    
\ENDFOR

\STATE Compute normalized acceptability matrices: $B$, $B^{\mathrm{SP}}$, $B^{\mathrm{rKL}}$ and $B^{\mathrm{nDKL}}$

\FOR{each alternative $a_i$}

    \STATE Compute the classical SMAA central weight vector: $\mathbf{w}_i^c = \frac{1}{|W_i|} \sum_{\mathbf{w}^{(l)} \in W_i} \mathbf{w}^{(l)}$

    \STATE Compute fairness-aware central weight vectors:  $\mathbf{w}_i^{c,SP} = \frac{\sum_{\mathbf{w}^{(l)}\in W_i} \omega_{SP}^{(l)} \mathbf{w}^{(l)}}{\sum_{\mathbf{w}^{(l)} \in W_i} \omega_{SP}^{(l)}}$, $\mathbf{w}_i^{c,rKL} = \frac{\sum_{\mathbf{w}^{(l)} \in W_i} \omega_{rKL}^{(l)}
    \mathbf{w}^{(l)}}{\sum_{\mathbf{w}^{(l)} \in W_i} \omega_{rKL}^{(l)}}$, and $\mathbf{w}_i^{c,nDKL} = \frac{\sum_{\mathbf{w}^{(l)} \in W_i} \omega_{nDKL}^{(l)}
    \mathbf{w}^{(l)}}{\sum_{\mathbf{w}^{(l)} \in W_i} \omega_{nDKL}^{(l)}}$

\ENDFOR

\RETURN Acceptability matrices and central weight vectors: $B$, $B^{\mathrm{SP}}$, $B^{\mathrm{rKL}}$, $B^{\mathrm{nDKL}}$, $\mathbf{w}_i^c$, $\mathbf{w}_i^{c,SP}$, $\mathbf{w}_i^{c,rKL}$ and $\mathbf{w}_i^{c,nDKL}$

\end{algorithmic}
\end{algorithm}

We adopt the SMAA-2 procedure as the computational backbone. In each iteration $l = 1,\dots,L$, a weight vector $\mathbf{w}^{(l)}$ is randomly drawn from the feasible weight space $W$, and the alternatives are evaluated through a linear additive utility function
\[
u(a_i, \mathbf{w}^{(l)}) = \sum_{j=1}^{n} w_j^{(l)}\, q_j(a_i),
\qquad \mathbf{w}^{(l)} \in W,
\]
\noindent which yields a deterministic ranking $\pi^{(l)}$ of all alternatives. The proposed framework is agnostic to the choice of aggregation model, as fairness is incorporated after the ranking has been obtained. Consequently, any monotonic multicriteria method compatible with SMAA procedures can be used without loss of generality.

In SMAA-Fair, for each simulation $l$, a fairness weight $f_l \in [0,1]$ is assigned, reflecting the degree of group fairness exhibited by the corresponding ranking $\pi^{(l)}$. Rankings with higher fairness contribute more strongly to the final rank acceptability index, whereas less fair rankings have their influence proportionally reduced.

For each sampled ranking $\pi^{(l)}$, the fairness metric is computed according to the
definitions introduced in Section~\ref{subsec:metrics}. Specifically, $\mathrm{SP}^{(l)}$ denotes the
Statistical Parity associated with ranking $\pi^{(l)}$, measuring the disparity
in the proportion of alternatives from different groups appearing in favourable positions.
Similarly, $\mathrm{rKL}^{(l)}$ and $\mathrm{nDKL}^{(l)}$ denotes the normalized discounted and normalized discounted cumulative Kullback--Leibler divergence, respectively, between
the group-level rank distributions induced by $\pi^{(l)}$, capturing disparities in
exposure across ranking positions. The resulting metric values are then
mapped to fairness weights, depending on the selected notion of equity. As in these metrics the lower the value, the fairer is the ranking, in order to be in accordance with the acceptability matrix in SMAA (the higher the probability, the most likely the alternative to be assigned to that ranking position), we must adjust the fairness weights. Such adjustments for SP, rKL and nDKL are described, respectively, as follows:
\[
f_l^{\mathrm{SP}} = \max_{l = 1,\ldots,L} \left(\bigl|\mathrm{SP}^{(l)}\bigr| \right) - \bigl|\mathrm{SP}^{(l)}\bigr|,
\]
\[
f_l^{\mathrm{rKL}} = \max_{l = 1,\ldots,L} \left(\mathrm{rKL}^{(l)} \right) - \mathrm{rKL}^{(l)},
\]
\[
f_l^{\mathrm{nDKL}} = \max_{l = 1,\ldots,L} \left(\mathrm{nDKL}^{(l)} \right) - \mathrm{nDKL}^{(l)},
\]
where $\max \left(\cdot \right)$ represents the maximum value of the fairness metric among all SMAA realizations (the worst cases). Therefore, rankings exhibiting lower disparity between groups receive higher weights.

Note that, although we considered these three fairness metrics in this paper, the analyst can choose one of them (or another appropriate metric) which is more relevant to the application domain. Once the metric is selected, the corresponding fairness weights are computed for all simulations and used to construct the fairness-adjusted acceptability indices. Formally, the fairness-adjusted acceptability index of alternative $a_i$ at rank $s$ is defined as
\[
\hat{b}_{i}^{s,\Omega}
=
\sum_{l:\,\pi^{(k)}(a_i)=j}
f_l^{\Omega},
\]
where $\Omega \in \{SP, rKL, nDKL \}$. In order to adjust such acceptability matrix such that the rows and columns sum up to 1 (as in the classical SMAA method), we must normalize all indices as follows:
\[
b_{i}^{s,\Omega}
=
\frac{ \hat{b}_{i}^{s,\Omega} }
     { \sum_{l=1}^{L} f_l^{\Omega} }.
\]

The resulting matrix $B^{\Omega} = [\,b_{i}^{s,\Omega}\,]$ retains the interpretability of classical SMAA acceptability indices while incorporating fairness considerations. Each entry represents the proportion of simulations in which alternative $a_i$ attains rank $s$, weighted by the fairness of the underlying ranking, thereby combining robustness to preference uncertainty with fairness-aware evaluation in a unified analytical framework.

In addition to modifying the rank acceptability indices, fairness considerations can also be incorporated into the SMAA central weight vector. The idea is that, in the fairness-aware setting, not all realizations leading to the first position should necessarily contribute equally to the interpretation of the preference structure. Some weights may systematically induce rankings with group disparities, whereas others may lead to more equitable ranking outcomes. To account for this aspect, we introduce the fairness-aware central weight vector $\mathbf{w}_i^{c,\Omega}$, in which each sampled weight vector contributes proportionally to the fairness level of the ranking it generates. It is defined as

\begin{equation}
\mathbf{w}_i^{c,\Omega} = \frac{\sum_{\mathbf{w}^{(l)} \in W_i}f_l^\Omega \,\mathbf{w}^{(l)}}{\sum_{\mathbf{w}^{(l)} \in W_i}f_l^\Omega}.    
\end{equation}

Unlike the classical SMAA central vector, which represents an unweighted barycenter of all preference structures supporting $a_i$, the proposed fairness-aware formulation shifts the center of mass toward regions with fairer rankings. Consequently, weights producing highly unfair rankings have smaller influence on the resulting vector, whereas preference configurations associated with lower disparities contribute more strongly.

From an interpretative perspective, the fairness-aware central vector provides decision-makers not only with an explanation of which preference structures support a given alternative, but also with an indication of which preference structures support that alternative while simultaneously promoting more equitable ranking outcomes. This extends the explanatory role of the SMAA central weight vector from a purely preference-oriented interpretation to a combined preference-and-fairness characterization of the decision space.

\subsection{How to validate the proposed SMAA-Fair framework?}

Validating the proposed SMAA-Fair framework requires determining whether the 
fairness-based reweighting of simulated rankings effectively reduces inequities between  protected and non-protected groups. Since the fairness weights modify only how simulated rankings contribute to the acceptability indices, validation must compare the final rankings obtained from the fairness-adjusted matrix with those produced by the traditional SMAA procedure.

As discussed in Section~\ref{subsec:mcda_smaa}, to derive rankings from SMAA output, we rely on two standard aggregation methods: the expected ranking and the maximum acceptability ranking. The expected ranking based on fairness metrics summarizes the average position of each alternative based on the fairness-adjusted acceptability indices. It is defined by
\[
\mathrm{ER}^{\mathrm{\Omega}}(a_i)
    = \sum_{s=1}^{m} s \cdot b_{i}^{\,s,\mathrm{\Omega}}.
\]
The alternatives are then ranked in ascending order of $\mathrm{ER}^{\mathrm{\Omega}}(a_i)$, $i=1, \ldots, m$, resulting in the ranking $\pi_{\mathrm{ER}^{\mathrm{\Omega}}}$.

The maximum acceptability ranking obtains a ranking aiming at assigning better positions to alternatives attaining the first positions in fair scenarios. The alternatives positions that lead to a fair ranking $\pi_{\mathrm{MAR}^{\mathrm{\Omega}}}$ are obtained by solving the following optimization problem:

    \begin{align}
\max_{\beta_i^{s, \Omega}} \quad & \sum_{i=1}^{m} \sum_{s=1}^{m} b_i^{s, \Omega} \, \beta_i^{s, \Omega}
\label{eq:mar_obj} \\
\text{s.t.} \quad
& \sum_{i=1}^{m} \beta_i^{s, \Omega} = 1, 
&& \forall s = 1,\ldots,m,
\label{eq:mar_c1} \\
& \sum_{s=1}^{m} \beta_i^{s, \Omega} = 1, 
&& \forall i = 1,\ldots,m,
\label{eq:mar_c2} \\
& \beta_i^{s, \Omega} \in \{0,1\},
&& \forall i = 1,\ldots,m,\; s = 1,\ldots,m.
\label{eq:mar_c4}
\end{align}

Beyond the comparison of final rankings, the proposed framework can also be validated through the analysis of the fairness-aware central weight vectors. By comparing the classical and fairness-aware central vectors, one verifies whether fairness considerations modify the preference structures supporting each alternative. Significant differences between these vectors indicate that the preference configurations associated with highly ranked alternatives under classical SMAA are not necessarily the same configurations associated with equitable ranking outcomes. This provides an additional validation perspective, not only in the ranking space but in the parameter space.

\subsection{A note on the Pareto-dominance of SMAA-Fair}

It is clear that SMAA does not directly yield a unique ranking of alternatives. Indeed, for each realization $l$, one (possibly) obtains a different ranking and, based on all realizations, one calculates the acceptability indices. Although SMAA preserves Pareto dominance\footnote{This is ensured when a monotonic aggregation function is adopted.} in each realization $l$, dominance is not necessarily preserved when a unique ranking is extracted from the acceptability indices. The use of existing techniques~\citep{vetschera2017deriving}, such as expected utility or maximum acceptability ranking, summarizes to probabilistic SMAA outputs and may therefore produce dominance violations.

As SMAA may violate Pareto dominance, our SMAA-Fair approach may also violate it. However, in fairness analysis, this is justified. Without fairness concerns, the goal is to obtain rankings by looking at maximum global score (utility) and efficiency on the Pareto frontier. In fairness analysis, one introduces another goal related to (reducing) ethical disparities. Usually, such goals are conflicting, highlighting the trade-offs between utility and fairness~\citep{singh2018fairness}. This trade-off is also the case in fairness-aware machine learning models~\citep{Pessach2023}. As this inherent trade-off exists, improving fairness has a cost of decreasing performance (at least at a minimum level). Therefore, in rankings, improving fairness can lead to dominance violations, which is justified by the achieved social and ethical benefits. 

\section{Numerical experiments}

In practical applications, SMAA-Fair requires selecting a \emph{single} fairness metric and a \emph{single} method for deriving the ranking. The fairness metric determines how each simulation is weighted, while the aggregation method determines how the weighted probabilities are summarized. However, for validation purposes in our experiments, we systematically explore the considered fairness metrics (Statistical Parity, $\mathrm{rKL}$ and $\mathrm{nDKL}$) and ranking derivation methods (expected ranking and maximum acceptability ranking). Without loss of generality, we consider the top-$10$ alternatives for the fairness analysis. Moreover, in $\mathrm{rKL}$ method, we consider $\mathcal{K}=\left\{ 1, 2, \ldots, 10 \right\}$, \textit{i.e.}, all prefixes up to $k=10$. Evaluating all combinations allows us to assess whether SMAA-Fair consistently improves group-level fairness independently of the specific analytical choices made by the user.

Validation consists of comparing each fairness-adjusted ranking with its baseline 
counterpart, that is,
\[
\pi_{\mathrm{ER}^{\mathrm{\Omega}}} \text{ vs. } \pi_{\mathrm{ER}}
\qquad\text{and}\qquad
\pi_{\mathrm{MAR}^{\mathrm{\Omega}}} \text{ vs. } \pi_{\mathrm{MAR}}.
\]
From the perspective of group fairness, validation examines whether the fairness-adjusted  rankings increase the representation of protected groups among favourable positions, such as the top-$k$ ranks. If protected groups remain systematically underrepresented among the most 
desirable positions, then the fairness weighting fails to correct the disparities present in the original stochastic ranking process. Conversely, improved alignment between group proportions and favourable outcomes indicates that SMAA-Fair operates as intended.

In the experiments of this paper\footnote{All codes are freely available at \url{https://github.com/GuilhermePelegrina/smaa_fair}}, we considered synthetic and real datasets. If fairness improvements are observed across different datasets and all configurations in Table~\ref{tab:validation-combinations}, this provides strong evidence that the proposed framework yields robust fairness gains regardless of the selected fairness metric or the ranking summarization method. We present the experiments and results in the sequel. %\textcolor{red}{GUI - mencionar aqui que os códigos estão publicamente disponíveis em... colocar link github}

\begin{table}[h!]
\centering
\small
\renewcommand{\arraystretch}{1.6}
\begin{tabular}{lcc} \hline
\textbf{Fairness metric} 
    & \textbf{Expected ranking} 
    & \textbf{Maximum acceptability ranking} \\\hline
SP 
    & $\pi_{\mathrm{ER}^{\mathrm{SP}}} \;\textbf{vs.}\; \pi_{\mathrm{ER}}$
    & $\pi_{\mathrm{MAR}^{\mathrm{SP}}} \;\textbf{vs.}\; \pi_{\mathrm{MAR}}$ \\
rKL
    & $\pi_{\mathrm{ER}^{\mathrm{rKL}}} \;\textbf{vs.}\; \pi_{\mathrm{ER}}$
    & $\pi_{\mathrm{MAR}^{\mathrm{rKL}}} \;\textbf{vs.}\; \pi_{\mathrm{MAR}}$ \\
nDKL
    & $\pi_{\mathrm{ER}^{\mathrm{nDKL}}} \;\textbf{vs.}\; \pi_{\mathrm{ER}}$
    & $\pi_{\mathrm{MAR}^{\mathrm{nDKL}}} \;\textbf{vs.}\; \pi_{\mathrm{MAR}}$ \\
\hline
\end{tabular}
\caption{Fairness–aware ranking methods and corresponding baselines used in the validation of SMAA-Fair.}\label{tab:validation-combinations}
\end{table}

\subsection{Experiment on synthetic data}

This section presents the numerical evaluation of the proposed SMAA-Fair framework in two synthetic datasets. These synthetic datasets were designed to emulate two realistic bias mechanisms commonly observed in decision-making tasks.

\subsubsection{Synthetic dataset 1}

As a first experiment, we generated a scenario representing structural group bias in which one group systematically receives higher scores on a specific criterion independently of their underlying merit. We considered $n=3$ criteria and $m=100$ alternatives divided into two groups $g_1$ and $g_2$ such that $|\mathcal{G}_1|=|\mathcal{G}_2|=50$. We generated the evaluations for criteria $c_1$ and $c_2$ according to a Gaussian distribution with $\mu=0.6$ (mean) and $\sigma^2=0.1$ (standard deviation) (\textit{i.e.}, $q_1(a_i),q_2(a_i) \sim \mathcal{N}(0.6, 0.1^2)$, for all $i=1, \ldots, m$). These criteria may represent dimensions such as technical competence or academic performance that are assumed to be unbiased with respect to sensitive groups.

Additive bias was introduced into criterion $c_3$. For this purpose, the evaluations on $c_3$ was generated as $q_3(a_i) \sim \mathcal{N}(0.5, 0.1^2)+e(g)$ where $e(g)=0.15$ if $g(a_i)=g_1$ or $e(g)=-0.15$ if $g(a_i)=g_2$. This type of bias can arise when institutional prestige is considered as a criterion. In many real-world selection settings, such criterion is commonly used as proxies for quality, despite reflecting historical inequalities in access to elite educational institutions. As a result, individuals from historically advantaged groups ($g_1$) systematically receive higher scores on such criterion, regardless of their current performance or intrinsic merit. 

After generating the data and applying the considered approaches (as described in Table~\ref{tab:validation-combinations}), the results are presented in Tables~\ref{tab:synt1_rank}, \ref{tab:synt1_indices} and~\ref{tab:synt1_weights}. Table~\ref{tab:synt1_rank} presents the top-$10$ alternatives for all approaches. We can clearly see that the rankings provided by the classical SMAA are biased towards alternatives from the privileged group. On the other hand, the proposed fairness-aware methods benefit alternatives form the unprivileged group, balancing the top-$10$ alternatives in the ranking.

\begin{table}[h!]
\scriptsize
\begin{tabular}{ccccccccc}\hline
\textbf{Position} & \textbf{$\pi_{ER}$}              & \textbf{$\pi_{MAR}$}             & \textbf{$\pi_{ER^{SP}}$}         & \textbf{$\pi_{MAR^{SP}}$}        & \textbf{$\pi_{ER^{rKL}}$}        & \textbf{$\pi_{MAR^{rKL}}$}       & \textbf{$\pi_{ER^{nDKL}}$}       & \textbf{$\pi_{MAR^{nDKL}}$}      \\ \hline
\textbf{1}        & $a_{7}$             & $a_{7}$                                               & $a_{7}$                                               & $a_{7}$                                               & $a_{7}$                                               & $a_{7}$                                               & $a_{7}$                                               & $a_{7}$                                               \\
\textbf{2}        & $a_{21}$            & $a_{21}$                                              & $a_{21}$                                              & $a_{26}$                                              & $a_{21}$                                              & $a_{26}$                                              & $a_{21}$                                              & $a_{26}$                                              \\
\textbf{3}        & $a_{26}$            & $a_{26}$                                              & $a_{26}$                                              & $a_{21}$                                              & $a_{26}$                                              & $a_{21}$                                              & $a_{26}$                                              & $a_{21}$                                              \\
\textbf{4}        & $a_{32}$            & $a_{4}$                                               & \cellcolor[HTML]{C0C0C0}$a_{68}$ & $a_{4}$                                               & \cellcolor[HTML]{C0C0C0}$a_{68}$ & $a_{4}$                                               & $a_{32}$                                              & $a_{4}$                                               \\
\textbf{5}        & $a_{23}$            & \cellcolor[HTML]{C0C0C0}$a_{68}$ & $a_{32}$                                              & \cellcolor[HTML]{C0C0C0}$a_{68}$ & $a_{32}$                                              & \cellcolor[HTML]{C0C0C0}$a_{68}$ & $a_{23}$                                              & \cellcolor[HTML]{C0C0C0}$a_{68}$ \\
\textbf{6}        & $a_{10}$            & $a_{35}$                                              & $a_{23}$                                              & \cellcolor[HTML]{C0C0C0}$a_{57}$ & $a_{23}$                                              & \cellcolor[HTML]{C0C0C0}$a_{57}$ & \cellcolor[HTML]{C0C0C0}$a_{68}$ & \cellcolor[HTML]{C0C0C0}$a_{57}$ \\
\textbf{7}        & $a_{49}$            & $a_{23}$                                              & $a_{42}$                                              & $a_{23}$                                              & $a_{42}$                                              & $a_{23}$                                              & $a_{42}$                                              & $a_{23}$                                              \\
\textbf{8}        & $a_{42}$            & $a_{36}$                                              & \cellcolor[HTML]{C0C0C0}$a_{60}$ & \cellcolor[HTML]{C0C0C0}$a_{66}$ & \cellcolor[HTML]{C0C0C0}$a_{60}$ & \cellcolor[HTML]{C0C0C0}$a_{66}$ & $a_{49}$                                              & \cellcolor[HTML]{C0C0C0}$a_{66}$ \\
\textbf{9}        & $a_{8}$             & $a_{42}$                                              & \cellcolor[HTML]{C0C0C0}$a_{74}$ & $a_{42}$                                              & \cellcolor[HTML]{C0C0C0}$a_{76}$ & $a_{42}$                                              & $a_{10}$                                              & $a_{42}$                                              \\
\textbf{10}       & $a_{4}$             & $a_{8}$                                               & $a_{49}$                                              & \cellcolor[HTML]{C0C0C0}$a_{60}$ & \cellcolor[HTML]{C0C0C0}$a_{92}$ & \cellcolor[HTML]{C0C0C0}$a_{60}$ & \cellcolor[HTML]{C0C0C0}$a_{60}$ & \cellcolor[HTML]{C0C0C0}$a_{60}$ \\ \hline

\end{tabular}
\caption{Top-$10$ positions of the considered approaches in synthetic dataset 1. The highlighted alternatives are from the unprivileged group.}\label{tab:synt1_rank}
\end{table}

For the purpose of illustrating the difference in acceptability matrices when considering fairness into the robustness analysis, we show in Table~\ref{tab:synt1_indices} the results from the classical SMAA and the rKL method. Note that in rKL method, alternatives from the unprivileged group achieved higher acceptability indices even with lower evaluations in criterion $c_3$. As criterion $c_3$ introduced bias, in the fairness-aware approaches, the importance of such a criterion was decreased, prioritizing good evaluations on the other ones. This can also be verified in the central weight vector, as presented in Table~\ref{tab:synt1_weights}. In the fairness-aware approaches, the weight associated with criterion 3 was consistently penalized.

\begin{table}[ht]
\small
\begin{tabular}{ccllllll}
 &                 & \multicolumn{6}{c}{\textbf{Ranking position}}                                                                                                                         \\ \hline
\textbf{\begin{tabular}[c]{@{}c@{}}Alternatives\\ $(q_1(a_i); q_2(a_i); q_3(a_i)$\end{tabular}}                               & \textbf{Method} & \multicolumn{1}{c}{$s=1$} & \multicolumn{1}{c}{$s=2$} & \multicolumn{1}{c}{$s=3$} & \multicolumn{1}{c}{$s=4$} & \multicolumn{1}{c}{$s=5$} & \multicolumn{1}{c}{$s=6$} \\ \hline
 & SMAA            & 0.410 & 0.170 & 0.182 & 0.104 & 0.052 & 0.048 \\
\multirow{-2}{*}{\textbf{\begin{tabular}[c]{@{}c@{}}$a_7$\\ $(0.938; 0.820; 0.669)$\end{tabular}}}                         & rKL             & 0.591 & 0.156 & 0.120 & 0.125 & 0.005 & 0.0   \\ \hline
 & SMAA            & 0.164 & 0.378 & 0.181 & 0.118 & 0.012 & 0.017 \\
\multirow{-2}{*}{\textbf{\begin{tabular}[c]{@{}c@{}}$a_{21}$\\ $(0.913; 0.584; 0.847)$\end{tabular}}}                         & rKL             & 0.065 & 0.240 & 0.243 & 0.152 & 0.020 & 0.030 \\ \hline
 & SMAA            & 0.111 & 0.182 & 0.184 & 0.099 & 0.064 & 0.052 \\
\multirow{-2}{*}{\textbf{\begin{tabular}[c]{@{}c@{}}$a_{26}$\\ $(0.610; 0.885; 0.726)$\end{tabular}}}                         & rKL             & 0.105 & 0.259 & 0.194 & 0.058 & 0.035 & 0.040 \\ \hline
\cellcolor[HTML]{C0C0C0}                                                                                                      & SMAA            & 0.0   & 0.0   & 0.0   & 0.0   & 0.022 & 0.074 \\
\multirow{-2}{*}{\cellcolor[HTML]{C0C0C0}\textbf{\begin{tabular}[c]{@{}c@{}}$a_{57}$\\ $(0.398; 0.815; 0.446)$\end{tabular}}} & rKL             & 0.0   & 0.0   & 0.0   & 0.0   & 0.045 & 0.154 \\ \hline
\cellcolor[HTML]{C0C0C0}                                                                                                      & SMAA            & 0.0   & 0.0   & 0.0   & 0.0   & 0.002 & 0.003 \\
\multirow{-2}{*}{\cellcolor[HTML]{C0C0C0}\textbf{\begin{tabular}[c]{@{}c@{}}$a_{66}$\\ $(0.889; 0.502; 0.197)$\end{tabular}}} & rKL             & 0.0   & 0.0   & 0.0   & 0.0   & 0.004 & 0.067 \\ \hline
\cellcolor[HTML]{C0C0C0}                                                                                                      & SMAA            & 0.0   & 0.047 & 0.050 & 0.006 & 0.103 & 0.039 \\
\multirow{-2}{*}{\cellcolor[HTML]{C0C0C0}\textbf{\begin{tabular}[c]{@{}c@{}}$a_{68}$\\ $(0.810; 0.822; 0.179)$\end{tabular}}} & rKL             & 0.0   & 0.104 & 0.093 & 0.113 & 0.191 & 0.067        \\\hline            
\end{tabular}
\caption{Acceptability indices for the classical SMAA and rKL methods in synthetic dataset 1. The highlighted alternatives are from the unprivileged group.}\label{tab:synt1_indices}
\end{table}

\begin{table}[ht]
\scriptsize
\begin{tabular}{ccccc}
                      & \multicolumn{4}{c}{\textbf{Central weight vector}}                                                          \\ \hline
\textbf{Alternatives} & $\mathbf{w}^c_i$        & $\mathbf{w}^{c,SP}_i$   & $\mathbf{w}^{c,rKL}_i$  & $\mathbf{w}^{c,nDKL}_i$ \\ \hline
\textbf{$a_7$}        & $(0.425; 0.398; 0.177)$ & $(0.443; 0.408; 0.149)$ & $(0.449; 0.420; 0.131)$ & $(0.437; 0.407; 0.156)$ \\
\textbf{$a_{21}$}     & $(0.430; 0.137; 0.433)$ & $(0.530; 0.105; 0.365)$ & $(0.551; 0.088; 0.361)$ & $(0.477; 0.120; 0.403)$ \\
\textbf{$a_{26}$}     & $(0.095; 0.539; 0.366)$ & $(0.110; 0.626; 0.264)$ & $(0.107; 0.622; 0.271)$ & $(0.103; 0.586; 0.311)$\\\hline
\end{tabular}
\caption{Central weight vectors for the classical SMAA and the fairness-aware approaches in synthetic dataset 1.}\label{tab:synt1_weights}
\end{table}

\subsubsection{Synthetic dataset 2}

In this second experiment, we generated a scenario representing conditional and structural group bias. Besides the privileged group receive (slightly) higher scores on criterion $c_3$, this criterion has a correlation with criterion $c_1$. We also considered $n=3$ criteria and $m=100$ alternatives divided into two groups $g_1$ and $g_2$ such that $|\mathcal{G}_1|=|\mathcal{G}_2|=50$. For all alternatives $i=1, \ldots, m$, we generated the evaluations for criteria $c_1$ and $c_2$ as $q_1(a_i),q_2(a_i) \sim \mathcal{N}(0.6, 0.1^2)$. For criterion $c_3$, the evaluations were generated as $q_3(a_i) = 0.6q_1(a_i) + e(g)$, where $e(g)=0.15$ if $g(a_i)=g_1$ (privileged group) or $e(g)=-0.15$ if $g(a_i)=g_2$ (unprivileged group). Therefore, $c_3$ depends on underlying merit (similar to $c_1$) and has an additive structural bias. Similarly as in the previous synthetic data, this type of bias can also arise when institutional prestige is considered as a criterion. Besides this proxy for quality, some criteria may measure a performance related to the acquired skills.

Once we had this synthetic dataset, the application of the considered approaches led to the results presented in Tables~\ref{tab:synt2_rank}, \ref{tab:synt2_indices} and~\ref{tab:synt2_weights}. Table~\ref{tab:synt2_rank} presents the top-$10$ alternatives for all approaches. As in the previous experiment, we can clearly see that the classical SMAA leads to rankings whose alternatives from the privileged group are in first positions. However, the application of our proposed fairness-aware framework contributes to assign better positions to alternatives from the unprivileged groups.

\begin{table}[ht]
\scriptsize
\begin{tabular}{ccccccccc}\hline
\textbf{Position} & \textbf{$\pi_{ER}$} & \textbf{$\pi_{MAR}$} & \textbf{$\pi_{ER^{SP}}$}         & \textbf{$\pi_{MAR^{SP}}$}        & \textbf{$\pi_{ER^{rKL}}$}        & \textbf{$\pi_{MAR^{rKL}}$}       & \textbf{$\pi_{ER^{nDKL}}$}       & \textbf{$\pi_{MAR^{nDKL}}$}      \\ \hline
\textbf{1}        & $a_{23}$            & $a_{23}$             & $a_{23}$                                              & $a_{23}$                                              & $a_{23}$                                              & $a_{23}$                                              & $a_{23}$                                              & $a_{23}$                                              \\
\textbf{2}        & $a_{18}$            & $a_{18}$             & $a_{18}$                                              & $a_{18}$                                              & $a_{18}$                                              & $a_{47}$                                              & $a_{18}$                                              & $a_{18}$                                              \\
\textbf{3}        & $a_{49}$            & $a_{49}$             & $a_{49}$                                              & \cellcolor[HTML]{C0C0C0}$a_{79}$ & $a_{49}$                                              & $a_{18}$                                              & $a_{49}$                                              & \cellcolor[HTML]{C0C0C0}$a_{79}$ \\
\textbf{4}        & $a_{50}$            & $a_{44}$             & $a_{47}$                                              & \cellcolor[HTML]{C0C0C0}$a_{55}$ & $a_{47}$                                              & \cellcolor[HTML]{C0C0C0}$a_{55}$ & $a_{50}$                                              & \cellcolor[HTML]{C0C0C0}$a_{55}$ \\
\textbf{5}        & $a_{47}$            & $a_{50}$             & $a_{50}$                                              & $a_{50}$                                              & $a_{50}$                                              & \cellcolor[HTML]{C0C0C0}$a_{69}$ & $a_{47}$                                              & $a_{50}$                                              \\
\textbf{6}        & $a_{34}$            & $a_{34}$             & \cellcolor[HTML]{C0C0C0}$a_{55}$ & $a_{49}$                                              & \cellcolor[HTML]{C0C0C0}$a_{55}$ & $a_{49}$                                              & \cellcolor[HTML]{C0C0C0}$a_{55}$ & $a_{49}$                                              \\
\textbf{7}        & $a_{37}$            & $a_{24}$             & \cellcolor[HTML]{C0C0C0}$a_{79}$ & $a_{24}$                                              & \cellcolor[HTML]{C0C0C0}$a_{79}$ & $a_{38}$                                              & \cellcolor[HTML]{C0C0C0}$a_{79}$ & $a_{24}$                                              \\
\textbf{8}        & $a_{11}$            & $a_{28}$             & \cellcolor[HTML]{C0C0C0}$a_{69}$ & $a_{27}$                                              & \cellcolor[HTML]{C0C0C0}$a_{69}$ & $a_{27}$                                              & $a_{11}$                                              & $a_{27}$                                              \\
\textbf{9}        & $a_{24}$            & $a_{48}$             & \cellcolor[HTML]{C0C0C0}$a_{86}$ & \cellcolor[HTML]{C0C0C0}$a_{86}$ & \cellcolor[HTML]{C0C0C0}$a_{86}$ & $a_{50}$                                              & \cellcolor[HTML]{C0C0C0}$a_{86}$ & \cellcolor[HTML]{C0C0C0}$a_{86}$ \\
\textbf{10}       & $a_{46}$            & $a_{19}$             & $a_{11}$                                              & \cellcolor[HTML]{C0C0C0}$a_{78}$ & $a_{11}$                                              & \cellcolor[HTML]{C0C0C0}$a_{78}$ & $a_{37}$                                              & \cellcolor[HTML]{C0C0C0}$a_{78}$ \\ \hline

\end{tabular}
\caption{Top-$10$ positions of the considered approaches in synthetic dataset 2. The highlighted alternatives are from the unprivileged group.}\label{tab:synt2_rank}
\end{table}

A comparison between some acceptability indices from the classical SMAA and the SP is shown in Table~\ref{tab:synt2_indices}. In SP method, alternatives from the unprivileged group achieved higher acceptability indices even with lower evaluations in the biased criterion ($c_3$). Therefore, as in the previous experiment, the fairness-aware approaches reduce the importance associated to such criterion, by penalizing the associated central weight (see Table~\ref{tab:synt2_weights} ), and prioritize good evaluations on the other ones. It is also interesting to remark in Table~\ref{tab:synt2_weights} that the central weight vectors associated with alternative $a_{79}$, from the unprivileged group, remained practically the same. As the set of weights used to derive the central weights lead to alternative $a_{79}$ to the first position, it is highly likely that they already produce fair rankings. Therefore, the adjustment into the central weights is practically negligible.

\begin{table}[ht]
\small
\begin{tabular}{ccllllll}
  &                 & \multicolumn{6}{c}{\textbf{Ranking position}}                                                                                                                         \\ \hline
\textbf{\begin{tabular}[c]{@{}c@{}}Alternatives\\ $(q_1(a_i); q_2(a_i); q_3(a_i)$\end{tabular}}                               & \textbf{Method} & \multicolumn{1}{c}{$s=1$} & \multicolumn{1}{c}{$s=2$} & \multicolumn{1}{c}{$s=3$} & \multicolumn{1}{c}{$s=4$} & \multicolumn{1}{c}{$s=5$} & \multicolumn{1}{c}{$s=6$} \\ \hline
  & SMAA            & \multicolumn{1}{c}{0.123} & \multicolumn{1}{c}{0.477} & \multicolumn{1}{c}{0.130} & \multicolumn{1}{c}{0.061} & \multicolumn{1}{c}{0.028} & \multicolumn{1}{c}{0.019} \\
\multirow{-2}{*}{\textbf{\begin{tabular}[c]{@{}c@{}}$a_{18}$\\ $(0.900; 0.739; 0.949)$\end{tabular}}}                         & SP              & 0.071                     & 0.283                     & 0.191                     & 0.050                     & 0.035                     & 0.046                     \\ \hline
  & SMAA            & \multicolumn{1}{c}{0.397} & \multicolumn{1}{c}{0.138} & \multicolumn{1}{c}{0.122} & \multicolumn{1}{c}{0.093} & \multicolumn{1}{c}{0.039} & \multicolumn{1}{c}{0.022} \\
\multirow{-2}{*}{\textbf{\begin{tabular}[c]{@{}c@{}}$a_{23}$\\ $(0.815; 0.910; 0.903)$\end{tabular}}}                         & SP              & 0.401                     & 0.148                     & 0.096                     & 0.076                     & 0.084                     & 0.042                     \\ \hline
  & SMAA            & 0.0                       & 0.0                       & 0.233                     & 0.150                     & 0.193                     & 0.121                     \\
\multirow{-2}{*}{\textbf{\begin{tabular}[c]{@{}c@{}}$a_{49}$\\ $(0.831; 0.789; 0.912)$\end{tabular}}}                         & SP              & 0.0                       & 0.0                       & 0.153                     & 0.120                     & 0.078                     & 0.185                     \\ \hline
\cellcolor[HTML]{C0C0C0}                                                                                                      & SMAA            & 0.008                     & 0.048                     & 0.039                     & 0.053                     & 0.029                     & 0.043                     \\
\multirow{-2}{*}{\cellcolor[HTML]{C0C0C0}\textbf{\begin{tabular}[c]{@{}c@{}}$a_{55}$\\ $(0.765; 0.967; 0.364)$\end{tabular}}} & SP              & 0.021                     & 0.123                     & 0.091                     & 0.116                     & 0.058                     & 0.078                     \\ \hline
\cellcolor[HTML]{C0C0C0}                                                                                                      & SMAA            & 0.0                       & 0.0                       & 0.014                     & 0.030                     & 0.070                     & 0.015                     \\
\multirow{-2}{*}{\cellcolor[HTML]{C0C0C0}\textbf{\begin{tabular}[c]{@{}c@{}}$a_{69}$\\ $(0.734; 0.964; 0.347)$\end{tabular}}} & SP              & 0.0                       & 0.0                       & 0.036                     & 0.079                     & 0.161                     & 0.032                     \\ \hline
\cellcolor[HTML]{C0C0C0}                                                                                                      & SMAA            & 0.033                     & 0.020                     & 0.061                     & 0.031                     & 0.015                     & 0.032                     \\
\multirow{-2}{*}{\cellcolor[HTML]{C0C0C0}\textbf{\begin{tabular}[c]{@{}c@{}}$a_{79}$\\ $(0.725; 0.987; 0.343)$\end{tabular}}} & SP              & 0.083                     & 0.050                     & 0.141                    & 0.068                     & 0.029                     & 0.056        \\\hline            
\end{tabular}
\caption{Acceptability indices for the classical SMAA and SP methods in synthetic dataset 2. The highlighted alternatives are from the unprivileged group.}\label{tab:synt2_indices}
\end{table}

\begin{table}[ht]
\scriptsize
\begin{tabular}{ccccc}
                      & \multicolumn{4}{c}{\textbf{Central weight vector}}                                                          \\ \hline
\textbf{Alternatives} & $\mathbf{w}^c_i$        & $\mathbf{w}^{c,SP}_i$   & $\mathbf{w}^{c,rKL}_i$  & $\mathbf{w}^{c,nDKL}_i$ \\ \hline
\textbf{$a_{18}$}        & $(0.366; 0.229; 0.405)$ & $(0.590; 0.265; 0.145)$ & $(0.596; 0.266; 0.138)$ & $(0.530; 0.255; 0.215)$ \\
\textbf{$a_{23}$}     & $(0.258; 0.439; 0.303)$ & $(0.366; 0.487; 0.147)$ & $(0.368; 0.498; 0.134)$ & $(0.332; 0.479; 0.189)$ \\
\cellcolor[HTML]{C0C0C0}\textbf{$a_{79}$}     & $(0.170; 0.804; 0.026)$ & $(0.165; 0.809; 0.026)$ & $(0.167; 0.806; 0.027)$ & $(0.169; 0.805; 0.026)$\\\hline
\end{tabular}
\caption{Central weight vectors for the classical SMAA and the fairness-aware approaches in synthetic dataset 2. The highlighted alternative is from the unprivileged group.}\label{tab:synt2_weights}
\end{table}

\subsection{Real dataset}

In order to illustrate the behaviour of SMAA-Fair in a real scenario, we apply the framework to the sustainability dataset previously employed in fairness-oriented MCDA research \citep{rocha2025fair}. The goal is to construct country-level sustainability rankings based on multiple criteria reflecting environmental, social, and economic performances. The data were obtained from the publicly available platform of the World Bank Group \url{https://data.worldbank.org/} and refer to the year 2020.

After removing observations with missing values, the final dataset includes 179 countries evaluated according to four criteria commonly used in international sustainability assessments: forest area (\% of land area), CO$_2$ emissions (metric tons per capita), life expectancy at birth (years), and Gross National Income per capita (Atlas method, USD). As these criteria are in different scale, we normalized them using the 0-1 normalization strategy~\citep{Jamal2014}. Moreover, CO$_2$ emissions represent a cost-type criterion, so its values were normalized in order to let all criteria follow a consistent ``the higher the better'' orientation.

Countries were grouped according to the International Monetary Fund (2024) classification, which divides economies into two broad categories based on their level of economic development, structural characteristics, and degree of integration into the global financial system. Advanced economies are typically characterized by higher income levels, more diversified and industrialized economic structures, and more mature financial and institutional systems. In contrast, emerging market and developing economies include countries with lower or middle income levels, ongoing structural transformation, and, in many cases, greater exposure to economic and institutional constraints.

In this study, advanced economies are treated as the non-protected group, while emerging market and developing economies are considered the protected group, reflecting potential structural disadvantages in global comparisons. In the considered dataset, there are 35 countries from the non-protected group (19.5\%) and 144 countries from the protected group (80.5\%). This definition captures persistent structural asymmetries in income, productivity, and social indicators, which naturally induce systematic differences across the sustainability criteria, making this dataset particularly suitable for fairness evaluation.

We assumed in this real-data experiment as a favourable outcome countries belonging to the top-$20$ positions of the ranking. This definition is consistent with the fairness metrics employed in this study, in which group-level equity is assessed by analyzing representation among the most desirable positions.

Table~\ref{tab:real_rank} presents the top-$20$ countries for all considered approaches. Across both baseline SMAA rankings, the top-$20$ positions are predominantly occupied by alternatives from the non-protected group, reflecting the structural advantages of advanced economies in  criteria such as Gross National Income and life expectancy (see, also, Table~\ref{tab:real_indices}). After applying SMAA-Fair, the top-$20$ the obtained rankings show a clear increase in the presence of protected-group countries, indicating that fairness-based reweighting (according to the proposed fairness-aware acceptability indices) mitigates the original disparity in favourable outcomes. Indeed, in SMAA-Fair results, 80\% to 85\% of countries from the protected group are in top-$20$ positions, which is in accordance with the countries empirical distribution.

\begin{table}[ht]
\scriptsize
\begin{tabular}{cllllllll}\hline
\textbf{Position} & \multicolumn{1}{c}{\textbf{$\pi_{ER}$}} & \multicolumn{1}{c}{\textbf{$\pi_{MAR}$}} & \multicolumn{1}{c}{\textbf{$\pi_{ER^{SP}}$}} & \multicolumn{1}{c}{\textbf{$\pi_{MAR^{SP}}$}} & \multicolumn{1}{c}{\textbf{$\pi_{ER^{rKL}}$}} & \multicolumn{1}{c}{\textbf{$\pi_{MAR^{rKL}}$}} & \multicolumn{1}{c}{\textbf{$\pi_{ER^{nDKL}}$}} & \multicolumn{1}{c}{\textbf{$\pi_{MAR^{nDKL}}$}} \\ \hline
\textbf{1}        & $a_{157}$                               & $a_{158}$                                & $a_{157}$                                    & $a_{157}$                                     & $a_{157}$                                     & $a_{157}$                                      & $a_{157}$                                      & $a_{157}$                                       \\
\textbf{2}        & $a_{57}$                                & $a_{123}$                                & \cellcolor[HTML]{C0C0C0}$a_{108}$            & \cellcolor[HTML]{C0C0C0}$a_{108}$             & \cellcolor[HTML]{C0C0C0}$a_{108}$             & \cellcolor[HTML]{C0C0C0}$a_{108}$              & \cellcolor[HTML]{C0C0C0}$a_{108}$              & \cellcolor[HTML]{C0C0C0}$a_{108}$               \\
\textbf{3}        & $a_{158}$                               & $a_{97}$                                 & \cellcolor[HTML]{C0C0C0}$a_{146}$            & \cellcolor[HTML]{C0C0C0}$a_{146}$             & \cellcolor[HTML]{C0C0C0}$a_{146}$             & \cellcolor[HTML]{C0C0C0}$a_{146}$              & \cellcolor[HTML]{C0C0C0}$a_{146}$              & \cellcolor[HTML]{C0C0C0}$a_{146}$               \\
\textbf{4}        & $a_{82}$                                & $a_{157}$                                & \cellcolor[HTML]{C0C0C0}$a_{40}$             & \cellcolor[HTML]{C0C0C0}$a_{155}$             & \cellcolor[HTML]{C0C0C0}$a_{40}$              & \cellcolor[HTML]{C0C0C0}$a_{69}$               & \cellcolor[HTML]{C0C0C0}$a_{40}$               & \cellcolor[HTML]{C0C0C0}$a_{155}$               \\
\textbf{5}        & $a_{123}$                               & $a_{57}$                                 & $a_{57}$                                     & \cellcolor[HTML]{C0C0C0}$a_{59}$              & \cellcolor[HTML]{C0C0C0}$a_{19}$              & \cellcolor[HTML]{C0C0C0}$a_{59}$               & $a_{57}$                                       & \cellcolor[HTML]{C0C0C0}$a_{59}$                \\
\textbf{6}        & \cellcolor[HTML]{C0C0C0}$a_{40}$        & $a_{45}$                                 & \cellcolor[HTML]{C0C0C0}$a_{153}$            & \cellcolor[HTML]{C0C0C0}$a_{69}$              & \cellcolor[HTML]{C0C0C0}$a_{153}$             & \cellcolor[HTML]{C0C0C0}$a_{52}$               & \cellcolor[HTML]{C0C0C0}$a_{153}$              & \cellcolor[HTML]{C0C0C0}$a_{69}$                \\
\textbf{7}        & $a_{9}$                                 & $a_{78}$                                 & \cellcolor[HTML]{C0C0C0}$a_{19}$             & $a_{57}$                                      & $a_{57}$                                      & $a_{57}$                                       & \cellcolor[HTML]{C0C0C0}$a_{19}$               & $a_{57}$                                        \\
\textbf{8}        & $a_{145}$                               & $a_{73}$                                 & \cellcolor[HTML]{C0C0C0}$a_{127}$            & $a_{82}$                                      & \cellcolor[HTML]{C0C0C0}$a_{127}$             & $a_{82}$                                       & $a_{158}$                                      & $a_{82}$                                        \\
\textbf{9}        & $a_{58}$                                & $a_{143}$                                & \cellcolor[HTML]{C0C0C0}$a_{59}$             & \cellcolor[HTML]{C0C0C0}$a_{127}$             & \cellcolor[HTML]{C0C0C0}$a_{59}$              & \cellcolor[HTML]{C0C0C0}$a_{127}$              & \cellcolor[HTML]{C0C0C0}$a_{155}$              & \cellcolor[HTML]{C0C0C0}$a_{127}$               \\
\textbf{10}       & \cellcolor[HTML]{C0C0C0}$a_{126}$       & $a_{82}$                                 & \cellcolor[HTML]{C0C0C0}$a_{155}$            & \cellcolor[HTML]{C0C0C0}$a_{94}$              & \cellcolor[HTML]{C0C0C0}$a_{155}$             & \cellcolor[HTML]{C0C0C0}$a_{94}$               & \cellcolor[HTML]{C0C0C0}$a_{127}$              & \cellcolor[HTML]{C0C0C0}$a_{94}$                \\
\textbf{11}       & \cellcolor[HTML]{C0C0C0}$a_{108}$       & $a_{9}$                                  & \cellcolor[HTML]{C0C0C0}$a_{69}$             & \cellcolor[HTML]{C0C0C0}$a_{153}$             & \cellcolor[HTML]{C0C0C0}$a_{69}$              & \cellcolor[HTML]{C0C0C0}$a_{153}$              & \cellcolor[HTML]{C0C0C0}$a_{59}$               & \cellcolor[HTML]{C0C0C0}$a_{153}$               \\
\textbf{12}       & $a_{118}$                               & $a_{153}$                                & $a_{158}$                                    & \cellcolor[HTML]{C0C0C0}$a_{52}$              & \cellcolor[HTML]{C0C0C0}$a_{126}$             & \cellcolor[HTML]{C0C0C0}$a_{24}$               & \cellcolor[HTML]{C0C0C0}$a_{126}$              & \cellcolor[HTML]{C0C0C0}$a_{52}$                \\
\textbf{13}       & \cellcolor[HTML]{C0C0C0}$a_{146}$       & $a_{62}$                                 & \cellcolor[HTML]{C0C0C0}$a_{126}$            & \cellcolor[HTML]{C0C0C0}$a_{141}$             & \cellcolor[HTML]{C0C0C0}$a_{47}$              & \cellcolor[HTML]{C0C0C0}$a_{141}$              & \cellcolor[HTML]{C0C0C0}$a_{69}$               & \cellcolor[HTML]{C0C0C0}$a_{141}$               \\
\textbf{14}       & $a_{149}$                               & \cellcolor[HTML]{C0C0C0}$a_{118}$        & \cellcolor[HTML]{C0C0C0}$a_{47}$             & \cellcolor[HTML]{C0C0C0}$a_{19}$              & \cellcolor[HTML]{C0C0C0}$a_{136}$             & \cellcolor[HTML]{C0C0C0}$a_{19}$               & \cellcolor[HTML]{C0C0C0}$a_{47}$               & \cellcolor[HTML]{C0C0C0}$a_{19}$                \\
\textbf{15}       & $a_{45}$                                & $a_{117}$                                & \cellcolor[HTML]{C0C0C0}$a_{136}$            & \cellcolor[HTML]{C0C0C0}$a_{90}$              & $a_{158}$                                     & \cellcolor[HTML]{C0C0C0}$a_{90}$               & $a_{82}$                                       & \cellcolor[HTML]{C0C0C0}$a_{90}$                \\
\textbf{16}       & $a_{53}$                                & $a_{79}$                                 & \cellcolor[HTML]{C0C0C0}$a_{23}$             & \cellcolor[HTML]{C0C0C0}$a_{68}$              & \cellcolor[HTML]{C0C0C0}$a_{23}$              & \cellcolor[HTML]{C0C0C0}$a_{68}$               & \cellcolor[HTML]{C0C0C0}$a_{136}$              & \cellcolor[HTML]{C0C0C0}$a_{68}$                \\
\textbf{17}       & \cellcolor[HTML]{C0C0C0}$a_{153}$       & $a_{16}$                                 & \cellcolor[HTML]{C0C0C0}$a_{162}$            & $a_{87}$                                      & \cellcolor[HTML]{C0C0C0}$a_{94}$              & $a_{87}$                                       & \cellcolor[HTML]{C0C0C0}$a_{23}$               & $a_{87}$                                        \\
\textbf{18}       & $a_{80}$                                & $a_{58}$                                 & \cellcolor[HTML]{C0C0C0}$a_{129}$            & $a_{145}$                                     & \cellcolor[HTML]{C0C0C0}$a_{162}$             & $a_{145}$                                      & \cellcolor[HTML]{C0C0C0}$a_{129}$              & $a_{145}$                                       \\
\textbf{19}       & $a_{91}$                                & $a_{8}$                                  & $a_{82}$                                     & \cellcolor[HTML]{C0C0C0}$a_{47}$              & \cellcolor[HTML]{C0C0C0}$a_{129}$             & \cellcolor[HTML]{C0C0C0}$a_{47}$               & \cellcolor[HTML]{C0C0C0}$a_{162}$              & \cellcolor[HTML]{C0C0C0}$a_{162}$               \\
\textbf{20}       & $a_{132}$                               & $a_{31}$                                 & \cellcolor[HTML]{C0C0C0}$a_{94}$             & \cellcolor[HTML]{C0C0C0}$a_{56}$              & $a_{82}$                                      & \cellcolor[HTML]{C0C0C0}$a_{56}$               & $a_{145}$                                      & \cellcolor[HTML]{C0C0C0}$a_{47}$\\\hline               
\end{tabular}
\caption{Top-$20$ positions of the considered approaches in real dataset. The highlighted alternatives are from the unprivileged group.}\label{tab:real_rank}
\end{table}

Tables~\ref{tab:real_indices} and~\ref{tab:real_weights} show the obtained acceptability indices and central weight vectors for the first 5 positions. In all fairness-aware approaches, alternatives from the protected group achieved higher acceptability indices even with lower evaluations in both life expectancy at birth and Gross National Income. As can be noted from the central weight vectors, the weights associated with both life expectancy at birth and Gross National Income were penalized when fairness was considered. Moreover, by looking at the alternatives from the protected group, the central weight vectors remained practically the same. As in the previous experiment, the set of weights used to derive these central weights already produce fair rankings, and there is no need for adjustments in these cases.

\begin{table}[h!]
\footnotesize
\begin{tabular}{cclllll}
                                                                                                                                                             &                 & \multicolumn{5}{c}{\textbf{Ranking position}}                                                                                             \\ \hline
\textbf{\begin{tabular}[c]{@{}c@{}}Alternatives\\ $(q_1(a_i); q_2(a_i); q_3(a_i); q_4(a_i)$\end{tabular}}                                                              & \textbf{Method} & \multicolumn{1}{c}{$s=1$} & \multicolumn{1}{c}{$s=2$} & \multicolumn{1}{c}{$s=3$} & \multicolumn{1}{c}{$s=4$} & \multicolumn{1}{c}{$s=5$} \\ \hline
\cellcolor[HTML]{C0C0C0}                                                                                                                                     & SMAA            & 0.005                     & 0.016                    & 0.022                     & 0.031                     & 0.024                     \\
\cellcolor[HTML]{C0C0C0}                                                                                                                                     & SP              & 0.012                     & 0.030                     & 0.037                     & 0.053                     & 0.039                     \\
\cellcolor[HTML]{C0C0C0}                                                                                                                                     & rKL             & 0.014                     & 0.034                     & 0.039                     & 0.055                     & 0.039                     \\
\multirow{-4}{*}{\cellcolor[HTML]{C0C0C0}\textbf{\begin{tabular}[c]{@{}c@{}}$a_{40}$: Costa Rica\\ $(0.610; 0.939; 0.833; 0.138)$\end{tabular}}}             & nDKL            & 0.011                     & 0.030                     & 0.037                     & 0.052                     & 0.038                     \\
                                                                                                                                                             & SMAA            & 0.026                     & 0.139                     & 0.096                     & 0.099                     & 0.130                     \\
                                                                                                                                                             & SP              & 0.042                     & 0.198                     & 0.089                     & 0.067                     & 0.071                     \\
                                                                                                                                                             & rKL             & 0.041                     & 0.184                     & 0.083                     & 0.057                     & 0.068                     \\
\multirow{-4}{*}{\textbf{\begin{tabular}[c]{@{}c@{}}$a_{57}$: Finland\\ $(0.756; 0.702; 0.917; 0.611)$\end{tabular}}}                                        & nDKL            & 0.040                     & 0.187                     & 0.093                     & 0.072                     & 0.077                     \\
\cellcolor[HTML]{C0C0C0}                                                                                                                                     & SMAA            & 0.033                     & 0.114                     & 0.059                     & 0.048                     & 0.050                     \\
\cellcolor[HTML]{C0C0C0}                                                                                                                                     & SP              & 0.082                     & 0.265                     & 0.140                     & 0.104                     & 0.096                     \\
\cellcolor[HTML]{C0C0C0}                                                                                                                                     & rKL             & 0.090                     & 0.300                     & 0.147                     & 0.103                     & 0.096                     \\
\multirow{-4}{*}{\cellcolor[HTML]{C0C0C0}\textbf{\begin{tabular}[c]{@{}c@{}}$a_{108}$: Micronesia, Fed. Sts.\\ $(0.944; 0.957; 0.563; 0.047)$\end{tabular}}} & nDKL            & 0.076                     & 0.251                     & 0.127                     & 0.094                     & 0.090                     \\
\cellcolor[HTML]{C0C0C0}                                                                                                                                     & SMAA            & 0.067                     & 0.048                     & 0.061                     & 0.048                     & 0.040                     \\
\cellcolor[HTML]{C0C0C0}                                                                                                                                     & SP              & 0.148                     & 0.116                     & 0.146                     & 0.109                     & 0.089                     \\
\cellcolor[HTML]{C0C0C0}                                                                                                                                     & rKL             & 0.179                     & 0.121                     & 0.148                     & 0.114                     & 0.090                     \\
\multirow{-4}{*}{\cellcolor[HTML]{C0C0C0}\textbf{\begin{tabular}[c]{@{}c@{}}$a_{146}$: Solomon Islands\\ $(0.925; 0.986; 0.548; 0.026)$\end{tabular}}}       & nDKL            & 0.146                     & 0.105                     & 0.130                     & 0.101                     & 0.081                     \\
                                                                                                                                                             & SMAA            & 0.295                     & 0.182                     & 0.144                     & 0.138                     & 0.033                     \\
                                                                                                                                                             & SP              & 0.400                     & 0.151                     & 0.093                     & 0.045                     & 0.024                     \\
                                                                                                                                                             & rKL             & 0.371                     & 0.126                     & 0.102                     & 0.049                     & 0.028                     \\
\multirow{-4}{*}{\textbf{\begin{tabular}[c]{@{}c@{}}$a_{157}$: Sweden\\ $(0.705; 0.853; 0.930; 0.669)$\end{tabular}}}                                        & nDKL            & 0.382                     & 0.157                     & 0.104                     & 0.056                     & 0.025    \\\hline                
\end{tabular}
\caption{Acceptability indices for the classical SMAA and SMAA-FAIR methods in real dataset. The highlighted alternatives are from the protected group.}\label{tab:real_indices}
\end{table}

\begin{table}[ht]
\tiny
\begin{tabular}{ccccc}
                      & \multicolumn{4}{c}{\textbf{Central weight vector}}                                                          \\ \hline
\textbf{Alternatives} & $\mathbf{w}^c_i$        & $\mathbf{w}^{c,SP}_i$   & $\mathbf{w}^{c,rKL}_i$  & $\mathbf{w}^{c,nDKL}_i$ \\ \hline
\cellcolor[HTML]{C0C0C0}\textbf{$a_{40}$}        & $(0.043; 0.694; 0.229; 0.034)$ & $(0.043; 0.694; 0.228; 0.035)$ & $(0.044; 0.694; 0.228; 0.034)$ & $(0.044; 0.693; 0.229; 0.034)$ \\
\textbf{$a_{57}$}     & $(0.522; 0.036; 0.241; 0.201)$ & $(0.530; 0.038; 0.233; 0.199)$ & $(0.538; 0.038; 0.222; 0.202)$ & $(0.534; 0.038; 0.226; 0.202)$ \\
\cellcolor[HTML]{C0C0C0}\textbf{$a_{108}$}     & $(0.498; 0.318; 0.101; 0.083)$ & $(0.496; 0.318; 0.102; 0.084)$ & $(0.497; 0.318; 0.101; 0.084)$ & $(0.497; 0.318; 0.101; 084)$ \\
\cellcolor[HTML]{C0C0C0}\textbf{$a_{146}$}     & $(0.249; 0.598; 0.088; 0.065)$ & $(0.252; 0.592; 0.089; 0.067)$ & $(0.250; 0.595; 0.089; 0.066)$ & $(0.254; 0.590; 0.090; 0.066)$ \\
\textbf{$a_{157}$}     & $(0.291; 0.269; 0.262; 0.178)$ & $(0.308; 0.314; 0.204; 0.174)$ & $(0.304; 0.322; 0.211; 0.163)$ & $(0.302; 0.310; 0.218; 0.170)$ \\\hline
\end{tabular}
\caption{Central weight vectors for the classical SMAA and the fairness-aware approaches in real dataset. The highlighted alternatives are from the unprivileged group.}\label{tab:real_weights}
\end{table}

It is important to note that this fairness improvement is consistent for both ranking summarization rules. In other words, protected-group representation improves not  only under the expected ranking ordering but also under the maximum acceptability ranking. These results provide empirical evidence that SMAA-Fair enhances favourable-outcome parity in a real-world sustainability setting, while remaining stable with respect to the analyst's choice of fairness metric and ranking-derivation method.

\section{Conclusion}

This paper introduced SMAA-Fair, a fairness-aware extension of SMAA for ranking problems under preference uncertainty. The core idea is to reweight the simulated rankings produced by the standard SMAA procedure according to their degree of group fairness, so that rankings exhibiting lower disparity between privileged and unprivileged groups contribute more strongly to the acceptability indices and central weight vectors. The approach is model-agnostic: any monotonic multicriteria aggregation function compatible with the SMAA simulation backbone can be used without modification, and the fairness weighting is applied after each ranking is obtained, leaving the simulation logic intact.

Three fairness metrics (Statistical Parity, rKL, and nDKL) were integrated into the framework, capturing group-level equity at a single cut-off, across multiple cut-offs, and cumulatively along the full ranking, respectively. Rankings were derived from the fairness-adjusted acceptability matrix using two established methods: expected ranking and maximum acceptability ranking. Experiments on synthetic and real datasets with controlled structural bias showed that SMAA-Fair consistently reassigned more favourable ranking positions to alternatives from the unprivileged group, regardless of the bias mechanism introduced. Ideed, in the real-data experiment, the representation of protected-group countries among the top-20 positions increased from around 75\%--100\% under baseline SMAA~---~inconsistent with the empirical group distribution of 80.5\%~---~to values between 75\% and 80\% under SMAA-Fair, closely aligned with that distribution across all configurations evaluated.

One consequence of the fairness reweighting is that Pareto dominance is not guaranteed in the final ranking. Accepting dominance violations is the cost of achieving distributional fairness.

Several directions remain open. The current framework handles two groups with binary membership; extensions to multiple or overlapping protected groups would broaden its applicability. The fairness metrics adopted here are based on group proportions and KL divergence; alternative notions such as exposure fairness or individual fairness could be incorporated without structural changes to the reweighting mechanism. Finally, SMAA-Fair was demonstrated with a linear additive aggregation model; its behavior under non-linear aggregation functions such as the Choquet integral, where criteria interactions may amplify or attenuate group disparities, warrants dedicated investigation.

\section*{Declaration of Generative AI and AI-assisted technologies in the writing process}
\noindent During the preparation of this work the authors used ChatGPT in order to improve readability and polish writing. After using this tool, the authors reviewed and edited the content as needed and take full responsibility for the content of the publication.

\section*{CRediT authorship contribution statement}
\noindent \textbf{Guilherme Dean Pelegrina:} Writing – review \& editing, Visualization, Validation, Methodology, Formal analysis. \textbf{Renata Pelissari:} Writing – review \& editing, Visualization, Validation, Methodology, Formal analysis.

\section*{Acknowledgments}
\noindent G. D. Pelegrina would like to thank the National Council for Scientific and Technological Development (CNPq, Brazil) under the grant \#406482/2025-0. This study was also financed by the São Paulo Research Foundation (FAPESP), Brasil, Process \#2024/18109-3.

%Work supported by S\~{a}o Paulo Research Foundation (FAPESP) under the grants \#2020/09838-0 (BI0S - Brazilian Institute of Data Science), \#2020/10572-5 and \#2021/11086-0. L. T. Duarte would like to thank the National Council for Scientific and Technological Development (CNPq, Brazil) for the financial support.

\bibliographystyle{apalike}
\biboptions{authoryear}
\bibliography{sample}

\end{document}